\documentclass[10pt, twocolumn]{article} 
\usepackage{newtxtext,newtxmath} 
\usepackage{lipsum} 
\usepackage[inner=1.5cm,outer=1.5cm,top=2cm, bottom=3.5cm]{geometry} 
\usepackage{graphicx} 
\usepackage{caption} 
\usepackage{authblk} 
\usepackage{soul} 
\usepackage{amsmath} 
\usepackage{tikzsymbols} 
\usepackage{dblfloatfix}
\usepackage{hyperref} 
\usepackage[switch]{lineno} 
\usepackage{gensymb}

\usepackage{subcaption}
\usepackage{graphicx}

\newif\ifdraft\draftfalse

\ifdraft

\newcommand\todo[1]{{\footnotesize \color{red}[#1 - \textbf{TODO}]}}

\newcommand\mpet[1]{{\footnotesize \color{blue}[#1 - \textbf{Maxime}]}}

\else
\newcommand\todo[1]{}
\newcommand\tb[1]{}
\newcommand\mpet[1]{}
\fi

\makeatletter 
\renewcommand\@maketitle{%
\hfill
\begin{minipage}{1\textwidth}
    \flushleft \vspace{-0.6cm}\scriptsize \textbf{Accepted at the 7th Int. Workshop on Image Analysis Methods in the Plant Sciences (IAMPS 2019). \hspace{5cm} 4-5 July 2019, Lyon, France}\\
    \centering
    \vspace{-0cm} 
    \let\footnote\thanks 
    {\LARGE \bf \@title \par }
    \vskip 0.5cm 
    {\large \bf \@author}
\end{minipage}
    \vskip 0cm \par
    \vspace{0.5cm} 
}

 \def\@oddfoot{\scriptsize \textbf{Accepted at the 7th Int. Workshop on Image Analysis Methods in the Plant Sciences (IAMPS 2019).\hfil 4-5 July 2019, Lyon, France}}%
 \def\@oddhead{\scriptsize \textbf{Accepted at the 7th Int. Workshop on Image Analysis Methods in the Plant Sciences (IAMPS 2019).\hfil 4-5 July 2019, Lyon, France}}%
 \def\@evenfoot{\scriptsize 4-5 July 2019, Lyon, France\hfil Accepted at IAMPS 2019}
 \def\@evenhead{\scriptsize 4-5 July 2019, Lyon, France\hfil Accepted at IAMPS 2019}
\makeatother

\renewenvironment{abstract}{
  \normalsize
  \begin{center}
  \bfseries 
  \abstractname
  \vspace{-0cm}\vspace{0pt} 
  \end{center}
  \list{}{
    \setlength{\leftmargin}{2cm}
    \setlength{\rightmargin}{\leftmargin}%
}%
\item\relax}
{\endlist}
 
\newenvironment{keywords}{
  \normalsize 
  \vspace{-1.2cm} 
  \begin{center}
  \end{center}
  \list{}{
    \setlength{\leftmargin}{2cm}
    \setlength{\rightmargin}{\leftmargin}%
  }%
  \item\relax
}
{\endlist}

\newenvironment{twocolsepvertical}{
  \vspace{-0.6cm} 
  \normalsize
  \begin{center}
  \end{center}
}

\newenvironment{body}{
  \normalsize
  \begin{center}
  \end{center}
  \vspace{-1.7cm} 
}

 \makeatletter
 \def\ps@headings{%
 \def\@oddhead{\mbox{}\scriptsize\rightmark \hfil \thepage}%
 \def\@evenhead{\scriptsize\thepage \hfil \leftmark\mbox{}}%
 \def\@oddfoot{\scriptsize \textbf{Preprint version. Accepted at the 7th International Workshop on Image Analysis Methods in the Plant Sciences (IAMPS 2019), July 2019.\hfil 4-5 July 2019, Lyon, France}}%
 \def\@evenfoot{\scriptsize 4-5 July 2019, Lyon, France\hfil Accepted at IAMPS 2019}}
 \makeatother

\begin{document}

\twocolumn
[{%
    \title{Toward a procedural fruit tree rendering framework for image analysis}
    \author[1]{Thomas Duboudin}
    \author[1]{Maxime Petit}
    \author[1]{Liming Chen}
    \affil[1]{LIRIS, CNRS UMR 5205, Ecole Centrale de Lyon, France}
    \date{} 
    \maketitle
    \vspace{-0.3cm}
    \begin{abstract}
    We propose a procedural fruit tree rendering framework, based on Blender and Python scripts allowing to generate quickly labeled dataset (\textit{i.e.} including ground truth semantic segmentation). It is designed to train image analysis deep learning methods (\textit{e.g.} in a robotic fruit harvesting context), where real labeled training datasets are usually scarce and existing synthetic ones are too specialized. Moreover, the framework includes the possibility to introduce parametrized variations in the model (\textit{e.g.} lightning conditions, background), producing a dataset with embedded \textit{Domain Randomization} aspect.
    \end{abstract}
    %
    \begin{twocolsepvertical}
    \end{twocolsepvertical}
    %
}]%

\begin{body}

\section{Introduction and Previous Work}

State-of-the-art methods for object recognition and grasping are currently mostly based on deep neural network. Despite tremendous results, these methods require a huge amount of labeled data in order to be trained. This is a major drawback in the field of robotics fruit-harvesting, where existing dataset are usually too small and/or  dedicated to a single specie (\textit{e.g.}~\cite{sa2016,barth2018a}).

We previously tackled this issue for indoor object grasping robots using a simulated environment and rendering engine~\cite{depierre2018, petit2018}. We apply here a similar strategy for fruits harvesting problems by defining a framework capable of generating scenes of fruit trees coupled with procedural scripts controlling parameters (\textit{e.g.} position of the fruits, type of background, lightning condition) to introduce realistic variations for outdoor data\footnote{Source code available at \href{https://github.com/tduboudi/IAMPS2019-Procedural-Fruit-Tree-Rendering-Framework}{https://github.com/tduboudi/IAMPS2019-Procedural-Fruit-Tree-Rendering-Framework}}.

One of the most photo-realistic synthetic dataset of fruit trees is the work of Barth \textit{et al.}~\cite{barth2018a} for sweet pepper. This come at the price of a huge computational cost (10 min/frame with a 16 core processor) preventing it to be easily extendable in order to produce large dataset for other fruits.  In fact, such high degree of photo-realism does not seem to be needed for deep simulated learning~\cite{rahnemoonfar2017deep}. That is why we aim at only an adequate photo-realism with quick rendering and easy-generation method, allowing scientists to create their own fruits or tree dataset according to their precise research interests.

\vspace{-0.2cm} 
\section{Material and Methods}

\begin{figure}

  \begin{subfigure}[t]{.32\linewidth}
    \centering
    \includegraphics[width=1.0\linewidth]{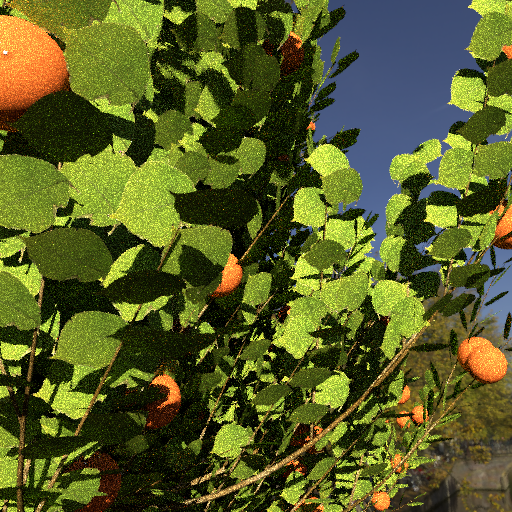}
  \end{subfigure}
  \hfill
  \begin{subfigure}[t]{.32\linewidth}
    \centering
    \includegraphics[width=1.0\linewidth]{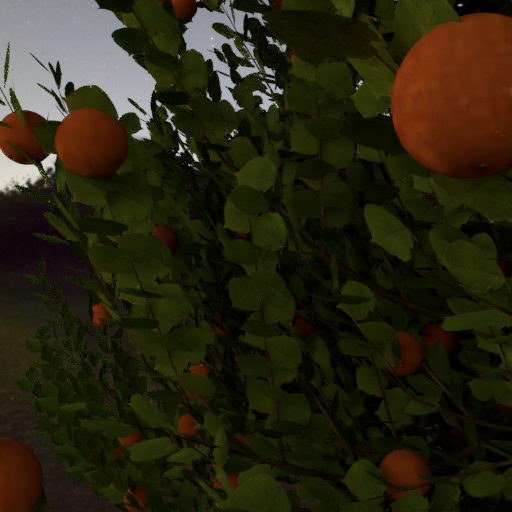}
  \end{subfigure}
  \hfill
  \begin{subfigure}[t]{.32\linewidth}
    \centering
    \includegraphics[width=1.0\linewidth]{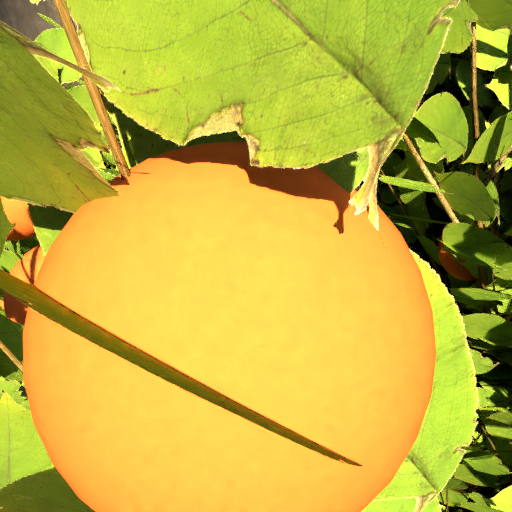}
  \end{subfigure}

  \medskip

  \begin{subfigure}[t]{.32\linewidth}
    \centering
    \includegraphics[width=1.0\linewidth]{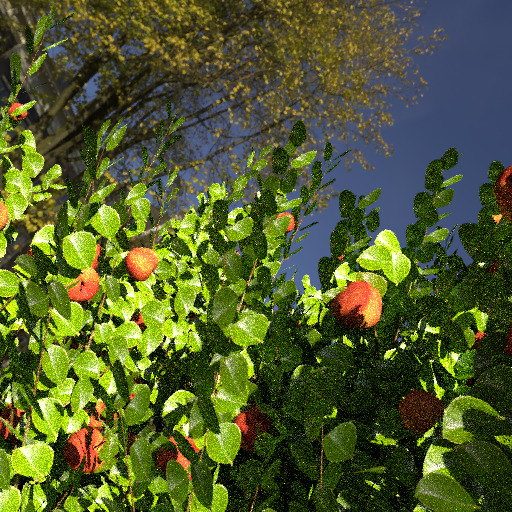}
  \end{subfigure}
  \hfill
  \begin{subfigure}[t]{.32\linewidth}
    \centering
    \includegraphics[width=1.0\linewidth]{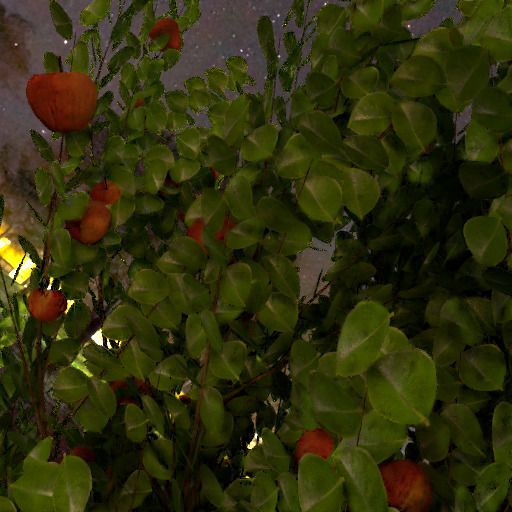}
  \end{subfigure}
  \hfill
  \begin{subfigure}[t]{.32\linewidth}
    \centering
    \includegraphics[width=1.0\linewidth]{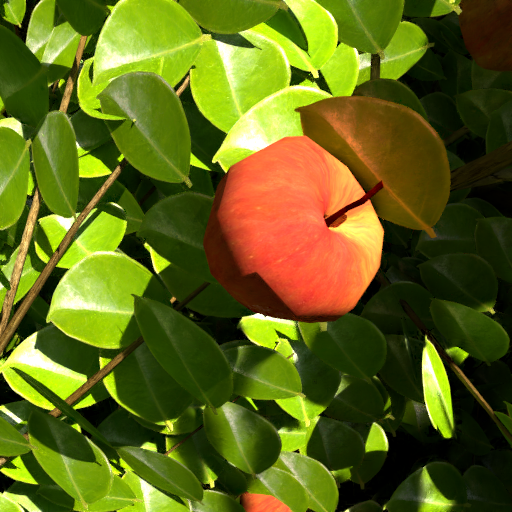}
  \end{subfigure}
  \caption{Samples of synthetic oranges (top) and apples (bottom) with different lightning conditions and rendering qualities.}
  \label{fig:front_page_figure}
\vspace{-0.5cm}  
\end{figure}


We chose the open-source Blender as the 3D-modelling and rendering software. Every options accessible through the Blender GUI can be reached and modified through a Python API, enabling us to entirely control the simulation with scripts.

The framework can be decomposed in two steps : first the generation of the tree or plant models, then a rendering script. It mainly contains a animation loop, such as that each time-step is responsible for the rendering of one image (and a semantic segmentation map, corresponding to the ground truth label), different from the previous ones. 

Tree models are generated following the rules of Weber and Penn~\cite{weber1995creation} using an existing Blender tree generation add-on, controlled by the Python generation script. It defines a number of parameters such as the branching frequency, the decrease in the radius of trunks and branches, the overall direction of the branches (up or down), the ratio tree height/branches length, \textit{etc}. The fruits and the leafs are randomly and uniformly added upon the naked tree following defined densities. 

The camera movements has to be defined in the rendering script, in which position and orientation of the camera should be directly provided at each time-step. We usually use a new random position and orientation at each iteration, such that the camera is globally pointing toward the models (see Fig.\ref{fig:camera_position}). While it has to be noted that the camera trajectories do not have to be continuous, we can also create robotic-like continuous trajectories toward branches on a tree. In addition, depth of field and rendering-engine parameters (\textit{e.g.} number of rays for the ray-tracing engine) can be randomly modified within a certain range every time-step. 

Background and lightning are also controlled in the rendering script, and can be easily changed every few time-steps to generate diverse images. This is a much needed feature allowing \textit{Domain Randomization}, known to be efficient in reducing the \textit{Reality Gap} encountered when transferring deep neural networks from simulation to the real world~\cite{tobin2017domain}. In order to have complex backgrounds we use freely available spherical HDRIs (high-dynamic-range 360\degree~images), which control the lightning and the background objects. The tree models are placed in the center of the HDRIs.

\vspace{-0.2cm} 
\section{Results and Discussion}

Overall, the rendering of one pair of images (512x512 pixels, raw and ground truth for semantic segmentation) is fairly quick with $\sim$[10s:30s] with GPU-rendering (NVIDIA GTX 1080). Fig.\ref{fig:front_page_figure} illustrates different possible variations for the scene generation (\textit{e.g.} fruits, light) and the quality of the rendering.

\begin{figure}[!t]
	\centering
    \includegraphics[width=0.85\linewidth]{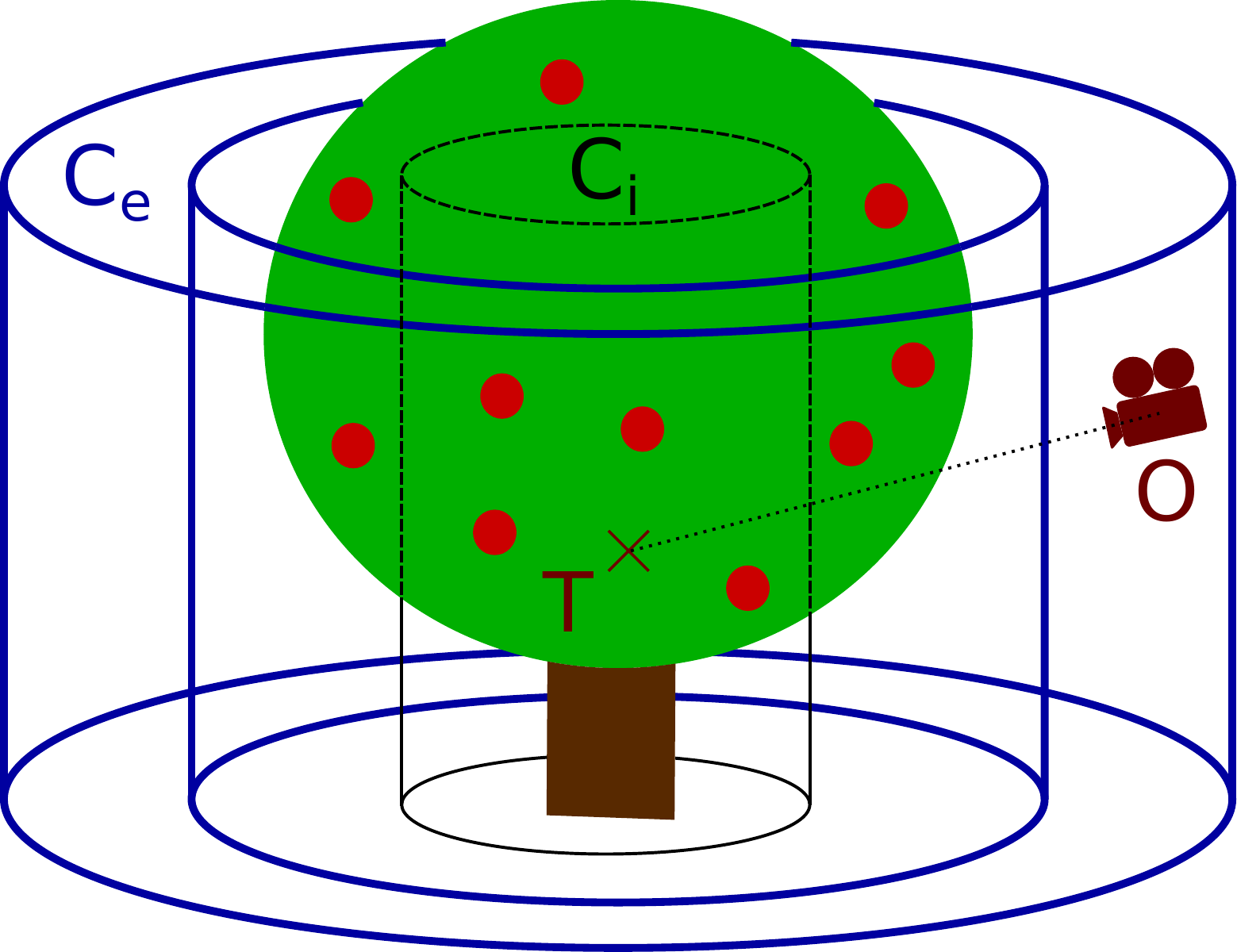}
	\caption{Schema of the camera point of view generation. Camera origin $O$  and target point $T$ are uniformly picked from respectively the hollow external cylinder $C_e$ and the internal cylinder $C_i$ (centered around the model).}
	\label{fig:camera_position}
	\vspace{-0.1cm} 
\end{figure}



\vspace{-0.2cm} 
\section{Conclusion and Future Work}

We designed a framework for semantic segmentation and object recognition field related to robotics fruit harvesting problems. It allows a quick and efficient fruit trees scene generation with parametrized variations, thus producing labelled images embedding a \textit{Domain Randomization} aspect.

We aim to extend our framework to reproduce the variations occurring during the lifetime of the fruits and tree. For instance, an implementation of the fruit senescence and decay~\cite{kider2011} unlocks the training of mature fruit picking robot and disease detection systems. We also plan to improve the photo-realism of the simulated image by using a Generative Adversarial Network as in ~\cite{barth2018b}, allowing to keep the rendering time needed quite low. 

\mpet{+deformable object if we have the space}

\vspace{-0.2cm} 
\section*{Acknowledgment}
This work is supported by the french National Research Agency (ANR), through the ARES labcom (grant ANR 16-LCV2-0012-01) and by the CHIST-ERA EU project "Learn-Real". 
\mpet{add the Learn-Real website when available}

\vspace{-0.2cm} 
\bibliographystyle{IEEEtran}
\bibliography{biblio} 

\begin{thebibliography}{1}
\providecommand{\url}[1]{#1}
\csname url@samestyle\endcsname
\providecommand{\newblock}{\relax}
\providecommand{\bibinfo}[2]{#2}
\providecommand{\BIBentrySTDinterwordspacing}{\spaceskip=0pt\relax}
\providecommand{\BIBentryALTinterwordstretchfactor}{4}
\providecommand{\BIBentryALTinterwordspacing}{\spaceskip=\fontdimen2\font plus
\BIBentryALTinterwordstretchfactor\fontdimen3\font minus
  \fontdimen4\font\relax}
\providecommand{\BIBforeignlanguage}[2]{{%
\expandafter\ifx\csname l@#1\endcsname\relax
\typeout{** WARNING: IEEEtran.bst: No hyphenation pattern has been}%
\typeout{** loaded for the language `#1'. Using the pattern for}%
\typeout{** the default language instead.}%
\else
\language=\csname l@#1\endcsname
\fi
#2}}
\providecommand{\BIBdecl}{\relax}
\BIBdecl

\bibitem{sa2016}
I.~Sa, Z.~Ge, F.~Dayoub, B.~Upcroft, T.~Perez, and C.~McCool, ``Deepfruits: A
  fruit detection system using deep neural networks,'' \emph{Sensors}, vol.~16,
  no.~8, p. 1222, 2016.

\bibitem{barth2018a}
R.~Barth, J.~IJsselmuiden, J.~Hemming, and E.~J. Van~Henten, ``Data synthesis
  methods for semantic segmentation in agriculture: A capsicum annuum
  dataset,'' \emph{Computers and electronics in agriculture}, vol. 144, pp.
  284--296, 2018.

\bibitem{depierre2018}
A.~Depierre, E.~Dellandr{\'e}a, and L.~Chen, ``Jacquard: A large scale dataset
  for robotic grasp detection,'' in \emph{2018 IEEE/RSJ Int. Conf. on
  Intelligent Robots and Systems (IROS)}, 2018, pp. 3511--3516.

\bibitem{petit2018}
M.~Petit, A.~Depierre, X.~Wang, E.~Dellandr{\'e}a, and L.~Chen, ``Developmental
  bayesian optimization of black-box with visual similarity-based transfer
  learning,'' in \emph{The 9th Joint IEEE Int. Conf. on Development and
  Learning and on Epigenetic Robotics (ICDL-Epirob)}, 2018.

\bibitem{rahnemoonfar2017deep}
M.~Rahnemoonfar and C.~Sheppard, ``Deep count: fruit counting based on deep
  simulated learning,'' \emph{Sensors}, vol.~17, no.~4, p. 905, 2017.

\bibitem{weber1995creation}
J.~Weber and J.~Penn, ``Creation and rendering of realistic trees,'' in
  \emph{Int. ACM Conf. on Computer graphics and interactive techniques}, 1995,
  pp. 119--128.

\bibitem{tobin2017domain}
J.~Tobin, R.~Fong, A.~Ray, J.~Schneider, W.~Zaremba, and P.~Abbeel, ``Domain
  randomization for transferring deep neural networks from simulation to the
  real world,'' in \emph{IEEE/RSJ Int. Conf. on Intelligent Robots and Systems
  (IROS)}, 2017, pp. 23--30.

\bibitem{kider2011}
J.~T. Kider~Jr, S.~Raja, and N.~I. Badler, ``Fruit senescence and decay
  simulation,'' in \emph{Computer Graphics Forum}, vol.~30, no.~2, 2011, pp.
  257--266.

\bibitem{barth2018b}
R.~Barth, J.~Hemming, and E.~J. van Henten, ``Improved part segmentation
  performance by optimising realism of synthetic images using cycle generative
  adversarial networks,'' \emph{arXiv preprint arXiv:1803.06301}, 2018.

\end{thebibliography}

\end{body}

\end{document}